\documentclass{article}

\usepackage{arxiv}

\usepackage[utf8]{inputenc} 
\usepackage[T1]{fontenc}    
\usepackage{hyperref}       
\usepackage{url}            
\usepackage{booktabs}       
\usepackage{amsfonts}       
\usepackage{nicefrac}       
\usepackage{microtype}      
\usepackage{lipsum}
\usepackage{graphicx}
\graphicspath{ {./images/} }

\title{AutoML-Based Drought Forecast with Meteorological Variables}

\author{
 Shiheng Duan \\
  Atmospheric Science Graduate Group\\
  University of California, Davis\\
  Davis, CA 95616 \\
  \texttt{shiduan@ucdavis.edu} \\
  \AND
  Xiurui Zhang\\
  Central South University\\
  Changsha, Hunan, 410008, China\\
  \texttt{showrain0000@gmail.com}
}

\begin{document}
\maketitle
\begin{abstract}
A precise forecast for droughts is of considerable value to scientific research, agriculture, and water resource management. With emerging developments of data-driven approaches for hydro-climate modeling, this paper investigates an AutoML-based framework to forecast droughts in the U.S. Compared with commonly-used temporal deep learning models, the AutoML model can achieve comparable performance with less training data and time. As deep learning models are becoming popular for Earth system modeling, this paper aims to bring more efforts to AutoML-based methods, and the use of them as benchmark baselines for more complex deep learning models. 
\end{abstract}


\section{Introduction}
Droughts are environmental disasters that have immense influence on agriculture, wildfire and water supplies \cite{hari2020increased}. 
There are various indicators to describe drought and its severity from different views, such as meteorological drought, socio-economic drought and ground water drought \cite{mishra2010review}. Drought monitor is an assessment of drought over U.S. that combines climate indices, numerical model outputs and domain expertise \cite{svoboda2002drought}. It classified drought into 5 categories as shown in Table \ref{tab:droughtmonitor}. It is also associated with other drought indices such as Palmer index, Standard Precipitation Index (SPI), and soil moisture. 

\begin{table}[h]
\caption{Drought monitor descriptions and its association with other drought indices. Adapted from \cite{svoboda2002drought}. }
\centering
\label{tab:droughtmonitor}
\begin{tabular}{ccccc}
\hline
Category & Description         & Soil Moisture & Palmer Drought & SPI          \\ \hline
D0       & Abnormally Dry      & 21 to 30      & -1.0 to -1.9   & -0.5 to -0.7 \\
D1       & Moderate Drought    & 11 to 20      & -2.0 to -2.9   & -0.8 to -1.2 \\
D2       & Severe Drought      & 6 to 10       & -3.0 to -3.9   & -1.3 to -1.5 \\
D3       & Extreme Drought     & 3 to 5        & -4.0 to -4.9   & -1.6 to -1.9 \\
D4       & Exceptional Drought & 0 to 2        & -5.0 or less   & -2.0 or less \\ \hline
\end{tabular}
\end{table}

The forecast or prediction of Earth's hydro-climate systems often involves process-based models and/or data-driven models. Process-based models are traditionally used to model Earth system components \cite{raleigh2012comparing, naabil2017water}. These models are designed based on governing equations, which are from our current understandings of Earth system. The biases are inevitable, either from approximations of these equations, or limitations of domain knowledge. Meanwhile, machine learning (ML) approaches are used to learn the processes directly from data. ML models have been applied to various environmental variables, such as air pollutants \cite{lv2022meteorology} and streamflow \cite{meng2019robust}. In addition to ML models, deep learning (DL) models are widely used in Earth system modeling due to the developments of Graphic Processing Units (GPUs). Especially for hydro-climate systems, they have achieved satisfying accuracy \cite{duan2020using, wunsch2022deep}. Comparing ML models and DL models, there are much less parameters in ML models and thus shorter training and tuning time, while with more trainable parameters and complex architectures, DL models often get better performance for non-linear tasks. 

For both the process-based models and data-driven models, parameter-tuning is necessary and important. The parameterizations in process-based models and hyperparameters in data-driven models largely affect model performance. The tuning process generally is a series of trail-and-error experiments, which requires lots of computational power to obtain satisfying models. These tuning burdens for ML and DL models can be reduced by automated machine learning, or AutoML \cite{he2021automl}. AutoML frameworks build data pipelines from original data to model evaluations via preprocessing, hyperparameter tuning and model generations. Although the computational power can not be saved (i.e., multiple candidate models are trained and validated), the domain expert do not need to spend much time on engineering such as ML and DL tunings and can focus more on scientific aspects. 

In this work, an automl framework is used to forecast drought with a benchmark dataset and its performance is compared against two mainstream time-series deep learning models. The paper is organized as following: Section \ref{sec:dataset} introduces the benchmark dataset and model architectures. The forecast results are in Section \ref{sec:result}. 

\section{Dataset and Models}\label{sec:dataset}
\subsection{DroughtED Benchmark}
The dataset is from DroughtED, which provides both drought monitors and associated meteorological variables across U.S. at county-level \cite{minixhofer2021droughted}. The drought monitor is available every Tuesday and meteorological variables are from NASA Prediction Of Worldwide Energy Resources
(POWER) project \cite{sparks2018nasapower}. In addition, the static parameters, such as slope, aspects, land use type, etc., are also included from Harmonized World Soil Database \cite{nachtergaele2010harmonized}. The task is formulated as the following: given the past 180-day meteorological variables and drought monitors, forecast the drought monitor in 1, 2, 3, 4, 5, 6 weeks ahead. The detailed input variables can be found in \cite{minixhofer2021droughted}. 

\subsection{AutoGluon}
AutoGluon is an AutoML framework that automates ML pipelines on tabular, text and image datasets \cite{erickson2020autogluon}. For environmental research, it has been applied to landslide hazards in \cite{qi2021autogluon}. Compared with other AutoML frameworks, AutoGluon doesn't emphasize hyperparamter tunings. Instead, it utilizes several base ML models (and some simple neural networks) to generate ensemble predictions, and the final prediction is a staking or multi-layer staking of outputs from base ML models. It also enables k-fold bagging strategies for each layer to increase its performance. The detailed training methods and benchmarks for AutoGluon is available in \cite{erickson2020autogluon}. In this work, we only use one-layer stacking and no bagging is used due to the limited computational resources. It is very likely that the performance will increase dramatically when enable these options. 

\section{Forecast Results}\label{sec:result}
The DroughtED benchmark compares F1 score and mean absolute errors (MAE) for Week1 to Week6 forecasts. In this paper, the model is trained in a different manner with the benchmark. Two models are trained: regression (for MAE comparisons) and classification (for F1 comparisons), while the benchmark DL models are trained with a regression loss and the classification performance is assessed with post-processing. In regression, the evaluation metrics is set to MAE, and macro-averaged F1 for classification. The drought monitor values in the dataset ranges from 0 to 5, with 0 representing no drought and 1 to 5 corresponding to D0 to D4 in Table \ref{tab:droughtmonitor}. Meteorological variables are normalized with their medians and inter quantile ranges and static features are normalized with their means and standard deviations. Training, validation and testing set are separated according to the benchmark \cite{minixhofer2021droughted}. The model is trained with 60\% data in the training set and validate in the rest of training set since the validation set in the benchmark only covers 1 year, spanning from 2010 to 2011. Thus, our comparison to the models from \cite{minixhofer2021droughted} is not completely fair. 

\begin{table}[h]
\caption{Model classification performance. LSTM+Transformer denotes the ensemble prediction from LSTM and Transformer. }
\label{tab:f1comparison}
\centering
\begin{tabular}{ccccccc}
\hline
                 & Week1 & Week2 & Week3 & Week4 & Week5 & Week6 \\ \hline
LSTM             & 0.811 & 0.723 & 0.639 & 0.568 & 0.505 & 0.475 \\
Transformer      & 0.670 & 0.629 & 0.603 & 0.564 & 0.506 & 0.467 \\
LSTM+Transformer & 0.836 & 0.727 & 0.651 & 0.582 & 0.516 & 0.480 \\
AutoGluon        & 0.837 & 0.696 & 0.619 & 0.560 & 0.512 & 0.479 \\ \hline
\end{tabular}
\end{table}

Due to the computational power limitation, only FastAI neural networks and lightGBM with extremely randomized trees are used in AutoGluon for classification. For regression, lightGBM, CatBoost, XGBoost and neural networks with FastAI and PyTorch are used. The classification and regression results on test set are listed in Table \ref{tab:f1comparison} and \ref{tab:maecomparison} respectively. Compared with the LSTM and Transformer models used in \cite{minixhofer2021droughted}, the AutoGluon model can achieve comparable performance for both regression and classification tasts. Specifically for Week1 forecast, the AutoGluon model obtain the highest F1 score and lowest MAE. All the models show decreasing performances with time, and the AutoGluon model achieves better performance for Week5 and Week6 than any single DL models for the classification task. Deep learning models indeed obtain better performance for Week2 to Week5, especially when evaluated with MAE. It probably indicates the nonlinearity for 2-5 week-lead forecasts, which requires the models' abilities to extract complex temporal features, while for Week1 forecast, there are less nonlinearity and thus AutoGluon models are more suitable. 

\begin{table}[h]
\caption{Model regression performance. F1 score from regression is by rounding the labels and outputs. }
\label{tab:maecomparison}
\centering
\begin{tabular}{ccccccc}
\hline
                 & Week1 & Week2 & Week3 & Week4 & Week5 & Week6 \\ \hline
MAE              &       &       &       &       &       &       \\ \hline
LSTM+Transformer & 0.135 & 0.198 & 0.254 & 0.321 & 0.388 & 0.427 \\
LSTM             & 0.178 & 0.237 & 0.265 & 0.328 & 0.395 & 0.433 \\
Transformer      & 0.159 & 0.215 & 0.267 & 0.335 & 0.398 & 0.435 \\
AutoGluon        & 0.111 & 0.232 & 0.315 & 0.366 & 0.410 & 0.455 \\ \hline
F1               &       &       &       &       &       &       \\ \hline
AutoGluon        & 0.791 & 0.662 & 0.594 & 0.498 & 0.466 & 0.431 \\ \hline
\end{tabular}
\end{table}

\section{Conclusions and Future Works}
This paper demonstrates the potential of AutoML-based drought forecast with the DroughtED dataset. Compared with LSTM and Transformer models, the AutoGluon model can achieve similar performance with less training samples and doesn't rely on GPUs. Especially for Wee1 forecast, the AutoGluon models surpass LSTM and Transformer. Since AutoGluon models are lightweight, it is beneficial to use them as references to DL models in practice, especially for researchers with limited access to computational resources. In addition, it is likely to continue increase the performance by using multi-layer stacking and cross-validation bagging strategies. These options will be investigated and assessed in future works. 

Besides the accuracy, interpretability and explainability are also important, especially for Earth sciences since the scope is to use these models to better understand the Earth system. There are lots of tree-based models in AutoGluon, which are less complex than neural networks. This enables us to better understand how and why the models make such predictions. Although there are lots of methods to interpret neural networks, they are not intuitive and the discrepancies between Earth science applications and traditional computer science tasks are not negligible (i.e., images with three color channels versus multiple 2-D meteorological fields). The preditability of drought monitors will be further analyzed along with the contribution from each predictor. Meantime, this paper aims to showcase the potential of AutoML applications, and suggests reconsideration of ML models over DL models. 

\bibliographystyle{unsrt}  
\bibliography{drought}  

\begin{thebibliography}{10}

\bibitem{hari2020increased}
Vittal Hari, Oldrich Rakovec, Yannis Markonis, Martin Hanel, and Rohini Kumar.
\newblock Increased future occurrences of the exceptional 2018--2019 central
  european drought under global warming.
\newblock {\em Scientific reports}, 10(1):1--10, 2020.

\bibitem{mishra2010review}
Ashok~K Mishra and Vijay~P Singh.
\newblock A review of drought concepts.
\newblock {\em Journal of hydrology}, 391(1-2):202--216, 2010.

\bibitem{svoboda2002drought}
Mark Svoboda, Doug LeComte, Mike Hayes, Richard Heim, Karin Gleason, Jim Angel,
  Brad Rippey, Rich Tinker, Mike Palecki, David Stooksbury, et~al.
\newblock The drought monitor.
\newblock {\em Bulletin of the American Meteorological Society},
  83(8):1181--1190, 2002.

\bibitem{raleigh2012comparing}
Mark~S Raleigh and Jessica~D Lundquist.
\newblock Comparing and combining swe estimates from the snow-17 model using
  prism and swe reconstruction.
\newblock {\em Water Resources Research}, 48(1), 2012.

\bibitem{naabil2017water}
Edward Naabil, BL~Lamptey, Joel Arnault, A~Olufayo, and Harald Kunstmann.
\newblock Water resources management using the wrf-hydro modelling system:
  Case-study of the tono dam in west africa.
\newblock {\em Journal of Hydrology: Regional Studies}, 12:196--209, 2017.

\bibitem{lv2022meteorology}
Yunqian Lv, Hezhong Tian, Lining Luo, Shuhan Liu, Xiaoxuan Bai, Hongyan Zhao,
  Shumin Lin, Shuang Zhao, Zhihui Guo, Yifei Xiao, et~al.
\newblock Meteorology-normalized variations of air quality during the covid-19
  lockdown in three chinese megacities.
\newblock {\em Atmospheric Pollution Research}, page 101452, 2022.

\bibitem{meng2019robust}
Erhao Meng, Shengzhi Huang, Qiang Huang, Wei Fang, Lianzhou Wu, and Lu~Wang.
\newblock A robust method for non-stationary streamflow prediction based on
  improved emd-svm model.
\newblock {\em Journal of hydrology}, 568:462--478, 2019.

\bibitem{duan2020using}
Shiheng Duan, Paul Ullrich, and Lele Shu.
\newblock Using convolutional neural networks for streamflow projection in
  california.
\newblock {\em Frontiers in Water}, 2:28, 2020.

\bibitem{wunsch2022deep}
Andreas Wunsch, Tanja Liesch, and Stefan Broda.
\newblock Deep learning shows declining groundwater levels in germany until
  2100 due to climate change.
\newblock {\em Nature communications}, 13(1):1--13, 2022.

\bibitem{he2021automl}
Xin He, Kaiyong Zhao, and Xiaowen Chu.
\newblock Automl: A survey of the state-of-the-art.
\newblock {\em Knowledge-Based Systems}, 212:106622, 2021.

\bibitem{minixhofer2021droughted}
Christoph Minixhofer, Mark Swan, Calum McMeekin, and Pavlos Andreadis.
\newblock Droughted: A dataset and methodology for drought forecasting spanning
  multiple climate zones.
\newblock In {\em Tackling Climate Change with Machine Learning: Workshop at
  ICML 2021}, 2021.

\bibitem{sparks2018nasapower}
Adam~H Sparks.
\newblock nasapower: a nasa power global meteorology, surface solar energy and
  climatology data client for r.
\newblock {\em Journal of Open Source Software}, 3(30):1035, 2018.

\bibitem{nachtergaele2010harmonized}
Freddy Nachtergaele, Harrij van Velthuizen, Luc Verelst, NH~Batjes, Koos
  Dijkshoorn, VWP van Engelen, Guenther Fischer, Arwyn Jones, and
  L~Montanarela.
\newblock The harmonized world soil database.
\newblock In {\em Proceedings of the 19th World Congress of Soil Science, Soil
  Solutions for a Changing World, Brisbane, Australia, 1-6 August 2010}, pages
  34--37, 2010.

\bibitem{erickson2020autogluon}
Nick Erickson, Jonas Mueller, Alexander Shirkov, Hang Zhang, Pedro Larroy,
  Mu~Li, and Alexander Smola.
\newblock Autogluon-tabular: Robust and accurate automl for structured data.
\newblock {\em arXiv preprint arXiv:2003.06505}, 2020.

\bibitem{qi2021autogluon}
Wenwen Qi, Chong Xu, and Xiwei Xu.
\newblock Autogluon: A revolutionary framework for landslide hazard analysis.
\newblock {\em Natural Hazards Research}, 1(3):103--108, 2021.

\end{thebibliography}


\end{document}